\documentclass{article} %
\PassOptionsToPackage{table}{xcolor}
\usepackage{iclr2027_conference,times}

\usepackage{amsmath,amsfonts,bm}

\def\eqref#1{equation~\ref{#1}}

\def\1{\bm{1}}

\DeclareMathAlphabet{\mathsfit}{\encodingdefault}{\sfdefault}{m}{sl}
\SetMathAlphabet{\mathsfit}{bold}{\encodingdefault}{\sfdefault}{bx}{n}

\usepackage{hyperref}
\usepackage{url}

\usepackage{booktabs}
\usepackage{multirow}
\usepackage{xspace}

\usepackage{mathtools}

\usepackage{booktabs} %

\usepackage[ruled]{algorithm2e} %
\usepackage{cleveref}
\usepackage{graphicx}

\usepackage{array} %
\usepackage{amssymb}  %
\usepackage{amsfonts} %
\usepackage{bm} 
\usepackage{xcolor}   %
\usepackage[table]{xcolor}   %
\usepackage{pifont}
\usepackage{makecell} %
\usepackage{graphicx} %
\usepackage{multirow} %
\usepackage{colortbl}
\usepackage{placeins}

\usepackage{tcolorbox}
\usepackage{enumitem}
\usepackage{xcolor}
\usepackage{subcaption}
\usepackage{wrapfig}

\usepackage{tcolorbox}
\usepackage{caption}
\usepackage{subcaption}
\definecolor{chart}{HTML}{1f77b4}

\newtcolorbox{example}[1][]{
  colback=chart!5!white,
  colframe=chart,
  floatplacement=tp*,  %
  title=\centering\bfseries\sffamily #1, %
  fonttitle=\bfseries\sffamily, %
  width=\textwidth, %
  nobeforeafter, %
}

\definecolor{amethyst}{rgb}{0.6, 0.4, 0.8}

\definecolor{lemon}{RGB}{255,247,0}
\definecolor{maize}{RGB}{250,237,94}
\definecolor{coco1}{HTML}{D9E4EC}
\definecolor{coco2}{HTML}{B7CFDC}
\definecolor{coco3}{HTML}{6AABD2}
\definecolor{coco4}{HTML}{385E72}
\hypersetup{
    colorlinks=true,
    linkcolor=coco3,
    filecolor=coco4,      
    urlcolor=coco3,
    citecolor=coco3,
}

\definecolor{mustard}{RGB}{255,219,89}
\definecolor{ocre}{RGB}{241,103,35}
\definecolor{Tangerine}{RGB}{253,128,8}
\definecolor{framegreen}{RGB}{153, 188, 133}
\definecolor{bggreen}{RGB}{235, 250, 228}
\definecolor{lightgray}{RGB}{239,240,241}
\definecolor{c0}{cmyk}{1,0.3968,0,0.2588} 
\definecolor{c1}{cmyk}{0,0.6175,0.8848,0.1490} 
\definecolor{c2}{cmyk}{0.1127,0.6690,0,0.4431} 
\definecolor{c3}{cmyk}{0.3081,0,0.7209,0.3255} 
\definecolor{c4}{RGB}{164, 16, 52}

\definecolor{g0}{HTML}{ffdbdc}
\definecolor{g1}{HTML}{ffefeb}
\definecolor{g2}{HTML}{FCFFEB}
\definecolor{g3}{HTML}{e9fcea}
\definecolor{g4}{HTML}{d4f8d4}

\definecolor{c0}{cmyk}{1,0.3968,0,0.2588} 
\definecolor{c1}{cmyk}{0,0.6175,0.8848,0.1490} 
\definecolor{c2}{cmyk}{0.1127,0.6690,0,0.4431} 
\definecolor{c3}{cmyk}{0.3081,0,0.7209,0.3255} 
\definecolor{c4}{RGB}{164, 16, 52}
\definecolor{orange}{HTML}{E66100}
\definecolor{bluex}{HTML}{0C7BDC}
\definecolor{yellow}{HTML}{FFC20A}
\definecolor{lightpurple}{HTML}{E6E6FA}
\definecolor{lightbluee}{HTML}{e8f4f8}
\definecolor{c5}{HTML}{EE4E4E}
\definecolor{gggggg}{HTML}{EFEFEF}
\newtcbox{\hlprimarytab}{on line, box align=base, colback=blue!12,colframe=white,size=fbox,arc=3pt, before upper=\strut, top=-2pt, bottom=-4pt, left=-2pt, right=-2pt, boxrule=0pt}
\newtcbox{\hlsecondarytab}{on line, box align=base, colback=orange!10,colframe=orange,size=fbox,arc=3pt, before upper=\strut, top=-2pt, bottom=-4pt, left=-2pt, right=-2pt, boxrule=0pt}
\newtcbox{\hlwhite}{on line, box align=base, colback=blue!12,colframe=white,size=fbox,arc=2pt, before upper=\strut, top=-2pt, bottom=-4pt, left=-2pt, right=-2pt, boxrule=0pt}
\newtcbox{\hlyellow}{on line, box align=base, colback=BlueGreen!10,colframe=white,size=fbox,arc=2pt, before upper=\strut, top=-2pt, bottom=-4pt, left=-2pt, right=-2pt, boxrule=0pt}

\definecolor{darkercheck}{RGB}{0, 128, 0}     %
\definecolor{darkercross}{RGB}{192, 0, 0}     %
\definecolor{mycolor_green}{HTML}{E6F8E0}
\definecolor{mycolor_blue}{HTML}{E7EFFA}
\definecolor{mycolor_orange}{HTML}{EEA851}
\definecolor{mycolor_deepblue}{HTML}{4396A4}
\definecolor{mycolor_gray}{HTML}{DAD7D7}
\definecolor{customBlue}{RGB}{42, 110, 177}

\title{\name{}: Generative Multi-Player World Model with Hierarchical Communication}

\author{
  \bf{Kaiyi Huang}\textsuperscript{$1$} \quad
  \bf{Wangbo Yu}\textsuperscript{$2{\dagger}$} \quad
  \bf{Yukun Huang}\textsuperscript{$2$} \quad
  \bf{Jiwen Yu}\textsuperscript{$1$} \quad
  \bf{Kunhao Liu}\textsuperscript{$2$} \\
  \bf{Donglin Yang}\textsuperscript{$1$} \quad
  \bf{Wenbo Hu}\textsuperscript{$2$} \quad
  \bf{Ying Shan}\textsuperscript{$2$} \quad
  \bf{Xihui Liu}\textsuperscript{$1{\dagger}$} \\
  \textsuperscript{1}The University of Hong Kong \quad
  \textsuperscript{2}ARC Lab, Tencent IEG \\
  \multicolumn{1}{@{}c@{}}{\makebox[\textwidth][c]{\url{https://karine-huang.github.io/Multics/}}}
}

\newcommand{\name}{\textit{Multics}}

\iclrfinalcopy %
\renewcommand{\headrulewidth}{0pt}
\begin{document}

\maketitle

\let\thefootnote\relax\footnotemark\footnotetext{$^{\dagger}$ Corresponding author.} 

\begin{figure}[ht]
    \centering
    \vspace{-3.0em}
        \includegraphics[width=\linewidth]{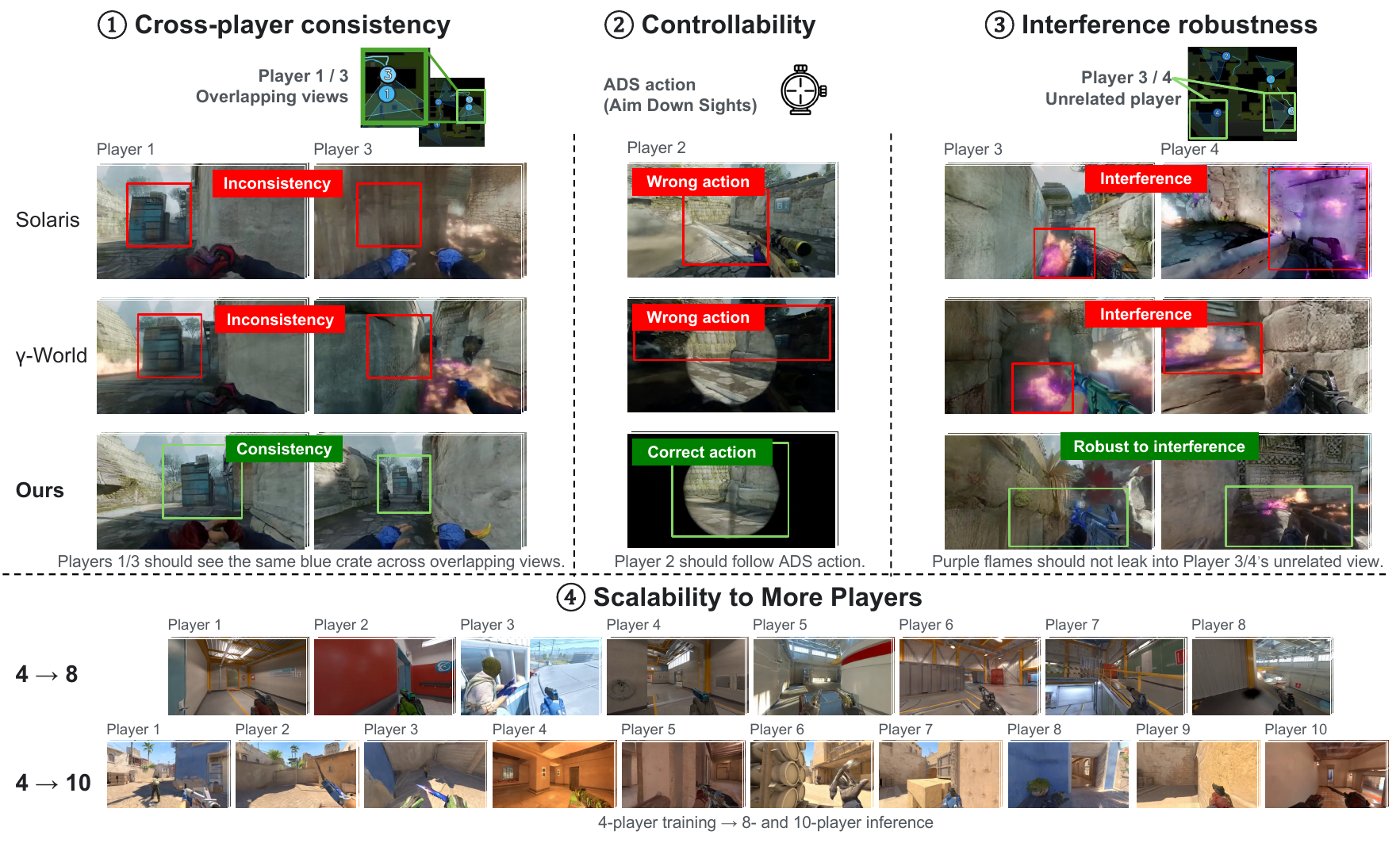}
        \vspace{-2.0em}
    \captionsetup{skip=7pt}
    \caption{
   \textbf{ ~\name{} is a multi-player video generation model with hierarchical communication that structures information exchange across player, group, and global levels.}
    ~\name{} can generate robust multi-player videos with cross-player consistency, player-specific controllability, and interference robustness, and can be generalized to more players without retraining.
    }
    \label{fig:teaser}
    \vspace{-7pt}
\end{figure}

\begingroup
\captionsetup[figure]{skip=2pt}
\captionsetup[table]{position=top,skip=2pt}
\setlength{\textfloatsep}{12pt}
\setlength{\dbltextfloatsep}{12pt}
\setlength{\floatsep}{12pt}
\setlength{\dblfloatsep}{12pt}
\setlength{\intextsep}{10pt}

\begin{abstract}
Multi-player interactive world models aim to simulate shared environments from multiple players' perspectives, but achieving robust generation while scaling to more players remains challenging. Existing approaches that rely on fully-connected communication (\textit{e.g.}, Solaris) incur quadratic attention costs as the number of players grows, while centralized approaches (\textit{e.g.}, $\gamma$-World) suffer from interference between unrelated events in a shared global representation.
To address these challenges, we present~\name{}, a multi-player video generation framework with hierarchical communication that structures information exchange across player, group, and global levels. 
This hierarchy introduces group tokens into self-attention as intermediate group-level representations that aggregate locally relevant player information.
These tokens connect to global tokens for cross-group communication, while the hierarchical topology limits direct mixing between unrelated local events.
Group-token parameters are shared across group slots, allowing the same communication module to accommodate varying numbers of players without introducing new group-specific parameters. 
We further propose budget-constrained dynamic grouping based on recent camera trajectories, which adaptively groups relevant players while limiting unnecessary communication. 
Experiments demonstrate that~\name{} achieves state-of-the-art performance in 4- and 6-player settings, with improved cross-player consistency, player-specific action and camera controllability, and interference robustness. Moreover,~\name{} generalizes from 4-player training to 6-, 8-, and 10-player inference without retraining, while reducing self-attention computation compared with fully-connected communication.
\end{abstract}

\maketitle
\section{Introduction}

Multi-player interactive video world models aim to simulate shared environments from multiple players' perspectives.
They have broad applications in game entertainment~\citep{bruce2024genie} and interactive simulation~\citep{ren2025cosmos, fung2025embodied}.
However, 
most video world models focus on single-player simulation~\citep{wang2026matrix3, hyworld2025, worldplay2025, lingbot-world} and generate observations conditioned on a single player's action, camera trajectory, and text prompt. 
Moving from single-player to multi-player simulation introduces two challenges: \textit{robust multi-player generation} and \textit{scalability to more players} (\Cref{fig:teaser}).
Robust multi-player generation requires generated observations to maintain cross-player consistency across overlapping views, faithfully follow player-specific action and camera controls, and remain robust to interference from unrelated concurrent events. Scalability to more players requires the model to generalize to larger player counts without retraining and scale efficiently as the number of players increases.

Existing multi-player video generation methods adopt different communication mechanisms for inter-player information exchange, but remain limited in jointly supporting robust generation and scalability. Fully-connected communication methods~\citep{savva2026solaris, mira2026} such as Solaris~\citep{savva2026solaris} enable unrestricted information exchange across players but incur quadratic computational cost as the player count grows. In contrast, centralized communication methods such as $\gamma$-World~\citep{liu2026gammaworld} improve computational scalability through a shared global representation. 
However, routing heterogeneous interactions through the same communication pathway can mix information from unrelated events, increasing cross-event interference.
This can further degrade cross-player consistency and player-specific controllability as concurrent interactions become more numerous and diverse.
These limitations motivate structured local and global communication under a limited communication budget.

To address these challenges, we present~\name{}, a multi-player video generation framework with hierarchical communication that structures information exchange across the player, group, and global levels. 
Our key insight is that players vary in their relevance to one another, motivating a hierarchy that combines group-level aggregation of locally relevant information with global communication across groups.
As shown in~\Cref{fig:pipeline}, 
the hierarchy incorporates group tokens into self-attention as intermediate group-level representations that aggregate locally relevant player information and connect to global tokens for cross-group communication.
This topology limits direct mixing between unrelated local events.
Group-token parameters are shared across group slots, allowing the same communication module to accommodate varying numbers of players without introducing new group-specific parameters. 
To determine which players should share group-level representations, we propose budget-constrained dynamic grouping based on recent camera trajectories. This strategy adapts group assignments to player relevance while limiting unnecessary communication.

Experimental results show that our method achieves state-of-the-art performance in 4- and 6-player settings, with strong cross-player consistency, controllability, and interference robustness. Models trained with 4 players further generalize to 6-, 8-, and 10-player inference without retraining.
Efficiency comparisons show substantially reduced self-attention computation relative to fully-connected communication.
Ablation studies verify the effectiveness of our key designs.
Our contributions can be summarized as follows:

\begin{itemize}
    \item We introduce~\name{}, a multi-player video generation framework with hierarchical communication that structures inter-player information exchange across player, group, and global levels.
    Group tokens provide intermediate group-level representations for local aggregation while preserving global communication for cross-group exchange.
    \item We propose shared group-token parameters across group slots, enabling the same communication module to accommodate varying numbers of players without introducing new group-specific parameters. We further introduce budget-constrained dynamic grouping based on recent camera trajectories to organize relevant players while limiting unnecessary communication.
    \item Extensive experiments demonstrate state-of-the-art performance in multi-player generation, with strong cross-player consistency, player-specific action and camera controllability, and interference robustness. 
    ~\name{} further demonstrates generalization to larger player counts without retraining and maintains computational efficiency.
\end{itemize}

\section{Related Work}

\textbf{Video Generation and Video World Models.}
Recent advances in video generation span Diffusion Transformers (DiTs)~\citep{peebles2023dit,sora}, next-token and masked-token prediction~\citep{villegas2022phenaki,chang2023muse,kondratyuk2023videopoet,yu2023magvit,yu2023language,chang2022maskgit}, and hybrid methods~\citep{chen2024diffusionforcing,deng2025autoregressive}, enabling high-quality video synthesis~\citep{Veo3,kling,kong2024hunyuanvideo,gen4,wan2025}. Conditioning on camera trajectories~\citep{bai2025recammaster,he2024cameractrl,wang2024motionctrl}, actions~\citep{yu2025gamefactory,valevski2025diffusion}, and multimodal references~\citep{MiniMaxh3} further enables controllable generation. Building on the world-model paradigm of predicting future states from past states and actions~\citep{ha2018recurrent}, video world models serve as visual simulators by generating future observations~\citep{bar2025navigation,liang2024dreamitate,yang2023unisim}. These models have been explored in robotics~\citep{li2026causal,gao2026dreamdojo,yang2023learning,ye2026world}, video games~\citep{fan2022minedojo,alonso2024diffusion,wang2026matrix3,valevski2025diffusion,xiao2026worldmem}, autonomous driving~\citep{ren2025cosmos,hu2023gaia}, and physical simulation~\citep{yuan2026inference}, demonstrating their potential as general-purpose visual simulators.

\textbf{Multi-Player World Models.}
Recent multi-player world models use different communication mechanisms to maintain cross-player consistency. Dense joint-generation methods~\citep{sun2026prismaworld,odyssey2026agora,hu2026wantofight}, including Solaris~\citep{savva2026solaris} and MIRA~\citep{mira2026}, couple player streams through shared or cross-agent attention, enabling fine-grained interactions with increasing computational cost as player counts grow. Centralized methods, including $\gamma$-World~\citep{liu2026gammaworld} and MultiWorld~\citep{wu2026multiworld}, improve scalability by aggregating information through sparse hubs or compact global visual representations~\citep{mo2026streaming}. Another line of work uses persistent or structured shared world states to decouple world-state evolution from per-player rendering~\citep{po2026multigen,cai2026mass,zhao2026population,mo2026streaming,huang2026programmable,ye2026worldcast}.
These shared-state methods primarily address population scalability through state-rendering decomposition.
Our work explores an orthogonal design dimension: scalable communication within learned visual representations, using a hierarchical player-group-global topology to structure information exchange across representation levels.

\textbf{Communication Topologies.}
Multi-agent communication includes dense exchange in CommNet~\citep{sukhbaatar2016learning}, selective or targeted communication in ATOC~\citep{jiang2018learning} and TarMAC~\citep{das2019tarmac}, and hierarchical exchange across agent groups~\citep{zhao2026hicomm}. While these methods primarily support multi-agent decision making, we study communication within high-dimensional video representations for multi-player world modeling.

\section{Method}
\label{method}

\begin{figure}[t]
    \centering
    \includegraphics[width=\linewidth]{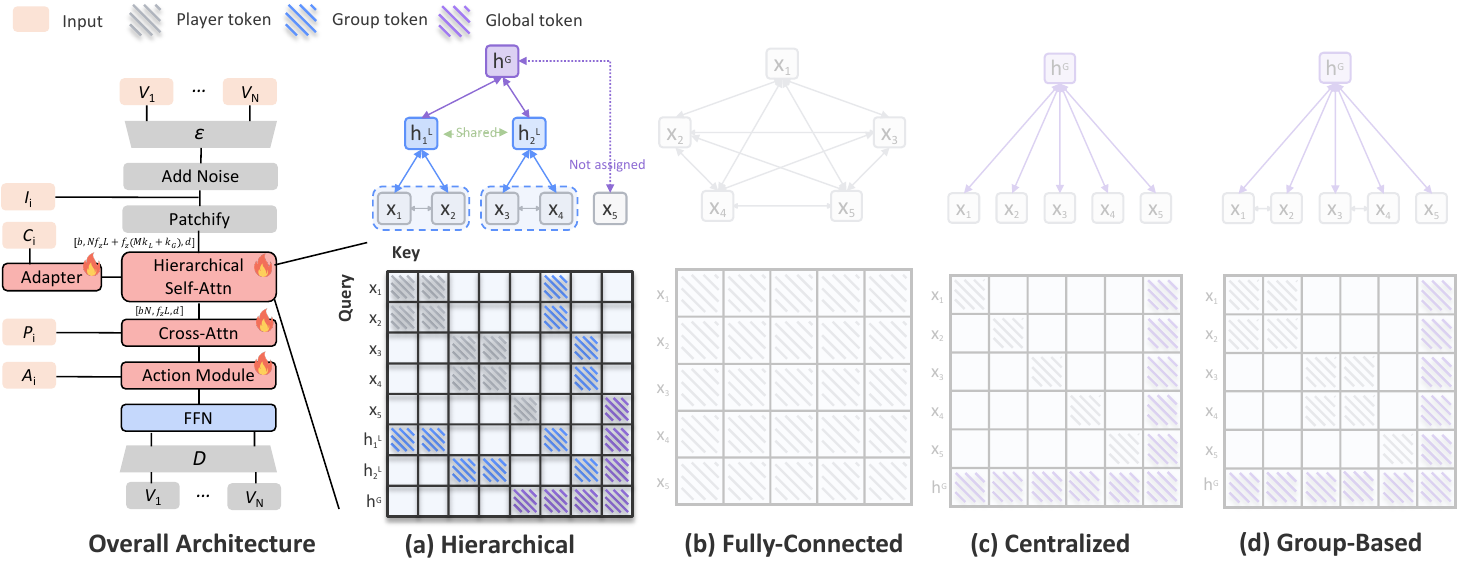}
    \caption{
    \textbf{Overview of~\name{}.} \textit{Left}: The training pipeline of~\name{}. The video diffusion model is trained to jointly generate multi-player videos $\mathcal{V}=\{V_i\}_{i=1}^{N}$, conditioned on each player $i$'s initial image $I_i$, action sequence $A_i$, camera trajectory $C_i$, and text prompt $P_i$.
    \textit{Right}: Comparison of different communication methods. 
    \textbf{(a)} Hierarchical communication in our method;
    \textbf{(b)} Fully-connected communication in~\citet{savva2026solaris};
    \textbf{(c)} Centralized communication in~\citet{liu2026gammaworld};
    \textbf{(d)} Group-based extension of centralized communication with direct intra-group player connections.
    The matrices illustrate attention connectivity at one latent timestep; block-causal temporal masking and spatial-token expansion are omitted for clarity.
    }

    \label{fig:pipeline}
\end{figure}

Given $N$ players in an environment, each player $i$ has an initial image $I_i \in \mathbb{R}^{c\times h\times w}$, a text prompt $P_i$, an action sequence $A_i \in \mathbb{R}^{(f-1)\times d_a}$, and a camera trajectory $C_i \in \mathbb{R}^{f\times3\times4}$, where $f,c,h,w,d_a$ denote the frame count, number of channels, height, width, and action dimension, respectively.
Our goal is to jointly generate videos $\mathcal{V}=\{V_i\}_{i=1}^{N}$, $V_i\in\mathbb{R}^{f\times c\times h\times w}$, conditioned on $\{I_i,P_i,A_i,C_i\}_{i=1}^{N}$.
The videos should maintain cross-player consistency when players observe overlapping views of the same scene or one another, and follow player-specific action and camera controls.
Thus, we introduce hierarchical communication across the player, group, and global levels, as shown in~\Cref{fig:pipeline}.

\subsection{Multi-Player Video Generation with Conditioning}
For each player $i$, the initial image $I_i$ serves as a clean anchor, while the action sequence $A_i$, camera trajectory $C_i$, and prompt $P_i$ condition the transformer blocks as detailed below.
All players' visual tokens are concatenated along the sequence dimension for inter-player communication (\Cref{subsection:hierarchical}).

\textbf{Action Conditioning.}
We decompose $A_i=(D_i,U_i)$ into discrete controls $D_i\in\mathbb{R}^{(f-1)\times23}$ and relative yaw/pitch changes $U_i\in\mathbb{R}^{(f-1)\times2}$~\citep{savva2026solaris}.
Following GameFactory~\citep{yu2025gamefactory}, we condition each latent frame on $rw_c$ action steps collected from a causal window spanning $w_c$ latent intervals. Here, $r$ is the video VAE's temporal compression ratio and $f_z=(f-1)/r+1$ is the latent frame count.
Discrete controls are projected to the transformer hidden dimension $d$, forming tokens in $\mathbb{R}^{f_z\times rw_c\times d}$ that are injected into the visual tokens via cross-attention.
Mouse controls are flattened into $\mathbb{R}^{f_z\times2rw_c}$ features and fused with visual tokens through a lightweight temporal module.

\textbf{Camera and Prompt Conditioning.}
Following ReCamMaster~\citep{bai2025recammaster}, we express all players' camera rotations and translations in a shared coordinate system anchored at the first player's initial frame, forming $\mathbf{C}\in\mathbb{R}^{N\times f\times3\times4}$.
Learnable camera adapters project $\mathbf{C}$ to the transformer hidden dimension and inject the resulting features into the corresponding player tokens' query and key features.
For the group and global tokens introduced in~\Cref{subsection:hierarchical}, we use zero camera features as placeholders~\citep{huang2026cinescene}.
Text prompts are encoded and injected via cross-attention~\citep{rombach2022high}.

\subsection{Hierarchical Communication for Multi-Player}
\label{subsection:hierarchical}
Effective multi-player generation requires structured information exchange across players.
Let $\mathbf{x}_i\in\mathbb{R}^{f_z\times L\times d}$ denote the visual
tokens of player $i$, where $L$ denotes the number of spatial tokens per frame.
All communication topologies below are implemented using self-attention masks that define the allowed token-to-token connections.
We review prior mechanisms~\citep{savva2026solaris,liu2026gammaworld} and their natural extensions before introducing our hierarchical design.

\textbf{Fully-Connected Communication.}
Solaris~\citep{savva2026solaris} concatenates the representations of all
players along the sequence dimension and performs shared self-attention,
allowing every player to communicate directly with all others:
\begin{equation}
    \mathbf{x}_{\mathrm{full}}
    =
    [\mathbf{x}_1,\mathbf{x}_2,\ldots,\mathbf{x}_N]_{\mathrm{seq}},
    \qquad
    \mathbf{x}_i \leftrightarrow \mathbf{x}_j,
    \quad i,j=1,\ldots,N.
\end{equation}
We denote this topology as ``fully-connected communication'' and illustrate
it in~\Cref{fig:pipeline}(b).

\textbf{Centralized Communication.}
Recent work~\citep{liu2026gammaworld} introduces shared global hub tokens
$\mathbf{h}^{G}\in\mathbb{R}^{f_z\times k_G\times d}$ as a common communication interface across players, where $k_G$ is the number of
global tokens per latent frame.
The communication sequence is
\begin{equation}
    \mathbf{x}_{\mathrm{cent}}
    =
    [\mathbf{x}_1,\ldots,\mathbf{x}_N,\mathbf{h}^{G}]_{\mathrm{seq}},
    \qquad
    \mathbf{x}_i \leftrightarrow \mathbf{h}^{G},
    \quad i=1,\ldots,N.
\end{equation}
We denote this topology as ``centralized communication'' and illustrate
it in~\Cref{fig:pipeline}(c).

\textbf{Group-Based Communication.}
As a straightforward extension, group-based communication introduces locality by restricting direct player-to-player communication to players within the same group.
Let $g(i)$ denote the group assignment of player $i$.
The resulting communication is
\begin{equation}
    \mathbf{x}_i \leftrightarrow \mathbf{x}_j
    \quad \mathrm{if}\quad g(i)=g(j),
    \qquad
    \mathbf{x}_i \leftrightarrow \mathbf{h}^{G}.
\end{equation}
We denote this topology as ``group-based communication'' and illustrate
it in~\Cref{fig:pipeline}(d).

\textbf{Hierarchical Communication (Ours).}
To improve cross-player consistency and player-specific controllability while mitigating interference among
unrelated local events, we introduce group tokens as intermediate group-level representations into self-attention.
Let $M$ denote the number of available group slots and
$g\in\{1,\ldots,M\}$ denote a group index.
For each group slot $g$, we instantiate $k_L$ group tokens
$\mathbf{h}^{L}_{g}\in\mathbb{R}^{f_z\times k_L\times d}$
between player- and global-level communication.
Let $\mathbf{h}^{L}$ denote the collection of group tokens from all $M$ group slots.
For player $i$ assigned to group $g(i)$, the hierarchical communication is
formulated as
\begin{equation}
    \mathbf{x}_{\mathrm{hier}}
    =
    [\mathbf{x}_1,\ldots,\mathbf{x}_N,
    \mathbf{h}^{L},\mathbf{h}^{G}]_{\mathrm{seq}},
    \qquad
    \mathbf{x}_i
    \leftrightarrow
    \mathbf{h}^{L}_{g(i)}
    \leftrightarrow
    \mathbf{h}^{G}.
\end{equation}
Players within the same group can directly attend to one another. 
Players that are not assigned to a group communicate directly through the global tokens.
We denote this topology as ``hierarchical communication'' and illustrate it
in~\Cref{fig:pipeline}(a).

To avoid binding group representations to fixed group identities,
we learn a set of shared group-token parameters
$\mathbf{E}^{L}\in\mathbb{R}^{k_L\times d}$.
At runtime, $\mathbf{E}^{L}$ is replicated across the $M$ group slots and
broadcast over latent frames to instantiate
$\mathbf{h}^{L}\in\mathbb{R}^{f_z\times M\times k_L\times d}$.
Although all group slots share the same learnable parameters, their hidden
representations evolve through hierarchical self-attention on the
players assigned to each group, yielding group-specific representations.
This design of shared group-token parameters makes the group slots exchangeable and enables the same communication module 
to accommodate varying numbers of players, without introducing new group-specific parameters or retraining the model. Appendix~\ref{appendix:shared_params} reports an ablation of shared group-token parameters.

\textbf{Comparison and Discussion.}
Our experimental results demonstrate the substantial advantages of hierarchical communication over fully-connected, centralized, and group-based alternatives, as illustrated in~\Cref{fig:qualitative-4_6players} and~\Cref{fig:group-based}.
We attribute these gains to the multi-granularity representations enabled by hierarchical communication. By introducing intermediate group-level representations between player- and global-level communication, our method allows locally related information to be aggregated before broader global exchange, reducing the direct mixing of heterogeneous information, such as actions, visual details, and unrelated local events, within a shared communication space. 
This facilitates more effective multi-player representation learning and improves cross-player consistency, player-specific action and camera controllability, and interference robustness.

\subsection{Dynamic Grouping}
\label{sec:dynamic_grouping}
Hierarchical communication requires determining which players should share group-level representations. 
Treating all players as equally relevant for group communication introduces unnecessary information exchange and computation and can increase cross-event interference. 
We therefore introduce dynamic grouping, which estimates pairwise player
relevance from recent camera trajectories~\citep{yu2025contextasmemory} and organizes relevant players into groups under a communication budget.

\textbf{Camera-Based Player Relevance.}
Building on the camera-based field-of-view (FoV) overlap in Context-as-Memory~\citep{yu2025contextasmemory}, we estimate player relevance from recent camera trajectories.
At each grouping update $t$, we compute a symmetric matrix $\mathbf{R}^{(t)}\in[0,1]^{N\times N}$ from observed camera histories: $R_{ij}^{(t)}=\operatorname{Rel}(C_i^{\leq t},C_j^{\leq t})$,
where $\operatorname{Rel}$ measures spatial proximity and viewing-direction alignment using only past and current camera states to preserve causality.
Higher scores indicate that the corresponding players are more likely to observe spatially related content and benefit from local exchange.
Relevance determines only the communication topology, while feature exchange is learned through hierarchical self-attention.
Details of computation are provided in Appendix~\ref{appendix:dynamic_grouping}.

\textbf{Budget-Constrained Dynamic Grouping.}
Given $N$ players, we determine the number of group slots $M$ from the player count and derive the maximum group capacity $K$ under a communication budget $\mathcal{C}_{\mathrm{max}}$ that matches the communication cost of the centralized baseline.
At time $t$, the active groups $\mathcal{G}^{(t)}$ must satisfy $|\mathcal{G}^{(t)}|\leq M$, $|\mathcal{S}|\leq K$ for all $\mathcal{S}\in\mathcal{G}^{(t)}$, and $\mathcal{C}(\mathcal{G}^{(t)})\leq\mathcal{C}_{\mathrm{max}}$, where $\mathcal{S}$ is a set of players assigned to one group and $\mathcal{C}$ measures the communication cost of the induced attention topology.
Using $\mathbf{R}^{(t)}$, we select a feasible grouping that maximizes player coverage while preserving high-relevance player relations.
Remaining group slots and budget are used to accommodate unassigned players; otherwise, they communicate only through global tokens.
Details of the determination of $M$, budget calculation, and derivation of $K$ are provided in Appendix~\ref{appendix:dynamic_grouping}.

\section{Experiments}

\subsection{Experiment Setting}

\textbf{Implementation Details.}
We train~\name{} based on Wan2.2 TI2V 5B~\citep{wan2025} using OpenCS2~\citep{blanchon2026opencs2}, an open multiplayer dataset rendered from ten-player Counter-Strike 2 matches with camera trajectories and mouse/keyboard action annotations.
We train separate models for $N\in\{4,6\}$, randomly selecting $N$ of the ten players for each sample.
Each sample contains $N$ videos of 165 frames at $352\times640$ resolution.
We adopt diffusion forcing training~\citep{chen2024diffusionforcing} and update the dynamic grouping topology every 48 video frames using causal camera trajectories.
Additional details are provided in Appendix~\ref{appendix:implementation}.

\textbf{Baselines.}
Given the lack of existing open-source methods for evaluating more than $4$ players, we adopt the following baselines.
(1) \textit{Fully-Connected Communication}: 
Solaris~\citep{savva2026solaris} pioneered multi-player video generation by concatenating all players' tokens along the sequence dimension for shared self-attention. 
(2) \textit{Centralized Communication}:
$\gamma$-World~\citep{liu2026gammaworld} introduces a set of shared hub tokens
for information aggregation across players.
For a fair comparison, we reimplement all baselines using the same base model and dataset, with identical training configurations and iterations.

\textbf{Evaluation Metrics.}
We employ 10 metrics across four key aspects: consistency, action accuracy, camera accuracy, and visual quality.
(1) \textit{Consistency}: following prior work~\citep{wang2024driving, liu2026openlongtail}, we evaluate cross-view consistency with GeoKPM, which uses geometry-aware keypoint matching under epipolar-geometry filtering.
Following~\citet{liu2026gammaworld}, we also report the pixel-aligned metrics PSNR and SSIM~\citep{wang2004ssim} and the perceptual metric LPIPS~\citep{zhang2018lpips}.
(2) \textit{Action accuracy}: following prior work~\citep{baker2022vpt, wu2026multiworld}, we use a trained Inverse Dynamics Model (IDM) to infer actions from generated videos.
(3) \textit{Camera accuracy}: we use RotErr, TransErr, and CamMC, following prior work~\citep{zheng2024cami2v,huang2026cinescene}.
(4) \textit{Visual quality}: following prior work~\citep{savva2026solaris,liu2026gammaworld}, we evaluate generation quality with FID~\citep{heusel2017fid} and FVD~\citep{unterthiner2018fvd}.
For evaluation, following~\citet{yu2025contextasmemory}, we randomly select 200 held-out OpenCS2 samples, with 100 for each separately trained model.
To assess cross-dataset applicability, we also train on EgoTeam's training set~\citep{peng2026egoteam}, a dataset of multi-robot egocentric videos rendered in reconstructed real-world indoor environments, and evaluate on 100 test samples.

\definecolor{sectiongray}{HTML}{F2F3F6}
\definecolor{oursblue}{HTML}{E1EFFA}

\begin{table*}[t]
    \centering
    \caption{
        \textbf{Quantitative comparison with previous methods}.
        We compare~\name{} with Solaris~\citep{savva2026solaris}
        and $\gamma$-World~\citep{liu2026gammaworld}
        on consistency, action accuracy, camera accuracy, and visual quality.
        \textit{Top}: Results with 4 and 6 players.
        \textit{Middle}: Generalization to more players without additional
        training, where $4 \rightarrow N$ denotes inference with $N$ players
        using models trained with 4 players.
        \textit{Bottom}: Results on EgoTeam.
        Bold numbers indicate the best result within each setting.
    }
    \label{tab:quantitative_comparison}
    \label{tab:generalization_egoteam}
    \setlength{\tabcolsep}{5pt}
    \renewcommand{\arraystretch}{1.0}
    \resizebox{\textwidth}{!}{%
    \begin{tabular}{lcccccccccc}
        \toprule[0.5pt]
        \multirow{2}{*}{\textbf{Method}}
        & \multicolumn{4}{c}{\textbf{Consistency}}
        & \multicolumn{1}{c}{\textbf{Action Accuracy}}
        & \multicolumn{3}{c}{\textbf{Camera Accuracy}}
        & \multicolumn{2}{c}{\textbf{Visual Quality}} \\
        \cmidrule(lr){2-5}
        \cmidrule(lr){6-6}
        \cmidrule(lr){7-9}
        \cmidrule(lr){10-11}
        & GeoKPM$\uparrow$
        & PSNR$\uparrow$
        & SSIM$\uparrow$
        & LPIPS$\downarrow$
        & IDM$\uparrow$
        & RotErr$\downarrow$
        & TransErr$\downarrow$
        & CamMC$\downarrow$
        & FID$\downarrow$
        & FVD$\downarrow$ \\
        \midrule

        \rowcolor{sectiongray}
        \multicolumn{11}{c}{
            \textbf{Top: Quantitative Comparison}
        } \\
        \midrule

        \multicolumn{11}{c}{\textit{4 Players}} \\
        \midrule
        Solaris~\citep{savva2026solaris}
        & 48.51\% & 17.62 & 0.4952 & 0.4324 & 93.87
        & 2.572 & 13.50 & 15.00 & 28.41 & 171.9 \\

        $\gamma$-World~\citep{liu2026gammaworld}
        & 49.42\% & 17.76 & \textbf{0.5009} & 0.4293 & 93.98
        & 2.728 & 13.66 & 15.26 & 29.36 & 200.5 \\

        \rowcolor{oursblue}
        \textbf{Ours}
        & \textbf{54.17\%} & \textbf{18.10} & 0.5005
        & \textbf{0.4184} & \textbf{94.17}
        & \textbf{2.434} & \textbf{11.67} & \textbf{13.21}
        & \textbf{26.39} & \textbf{162.3} \\
        \midrule

        \multicolumn{11}{c}{\textit{6 Players}} \\
        \midrule
        Solaris~\citep{savva2026solaris}
        & 54.65\% & 17.11 & \textbf{0.4728} & 0.4467 & 94.32
        & 3.991 & 11.38 & 14.19 & 18.83 & 115.1 \\

        $\gamma$-World~\citep{liu2026gammaworld}
        & 52.80\% & 17.21 & 0.4721 & 0.4389 & \textbf{94.34}
        & 3.761 & 11.59 & 14.29 & 20.16 & 128.4 \\

        \rowcolor{oursblue}
        \textbf{Ours}
        & \textbf{56.84\%} & \textbf{17.33} & 0.4687
        & \textbf{0.4311} & 94.33
        & \textbf{3.669} & \textbf{10.49} & \textbf{13.21}
        & \textbf{16.34} & \textbf{101.2} \\
        \midrule

        \rowcolor{sectiongray}
        \multicolumn{11}{c}{
            \textbf{Middle: Generalization to More Players}
        } \\
        \midrule

        \multicolumn{11}{c}{\textit{4 $\rightarrow$ 6 Players}} \\
        \midrule
        Solaris~\citep{savva2026solaris}
        & 49.18\% & 16.58 & 0.4674 & 0.4689 & 94.17
        & 3.979 & 12.22 & 14.86 & 23.16 & 148.2 \\

        $\gamma$-World~\citep{liu2026gammaworld}
        & 48.50\% & 16.74 & \textbf{0.4695} & 0.4618 & 94.20
        & 4.017 & 12.87 & 15.58 & 24.90 & 184.3 \\

        \rowcolor{oursblue}
        \textbf{Ours}
        & \textbf{54.33\%} & \textbf{16.97} & 0.4659
        & \textbf{0.4517} & \textbf{94.31}
        & \textbf{3.760} & \textbf{11.15} & \textbf{13.74}
        & \textbf{21.80} & \textbf{133.8} \\
        \midrule

        \multicolumn{11}{c}{\textit{4 $\rightarrow$ 8 Players}} \\
        \midrule
        Solaris~\citep{savva2026solaris}
        & 50.37\% & 16.38 & 0.4703 & 0.4778 & 94.16
        & 3.932 & 12.40 & 15.08 & 21.47 & 133.8 \\

        $\gamma$-World~\citep{liu2026gammaworld}
        & 49.17\% & 16.65 & \textbf{0.4712} & 0.4664 & 94.19
        & 3.794 & 11.75 & 14.53 & 23.45 & 172.3 \\

        \rowcolor{oursblue}
        \textbf{Ours}
        & \textbf{51.05\%} & \textbf{16.82} & 0.4676
        & \textbf{0.4600} & \textbf{94.26}
        & \textbf{3.553} & \textbf{11.52} & \textbf{13.92}
        & \textbf{20.07} & \textbf{117.0} \\
        \midrule

        \multicolumn{11}{c}{\textit{4 $\rightarrow$ 10 Players}} \\
        \midrule
        Solaris~\citep{savva2026solaris}
        & 47.82\% & 16.06 & \textbf{0.4670} & 0.4921 & 93.95
        & 7.320 & 12.91 & 18.68 & 21.28 & 139.0 \\

        $\gamma$-World~\citep{liu2026gammaworld}
        & 49.12\% & 16.38 & 0.4667 & 0.4741 & 94.11
        & 7.138 & 12.26 & 17.95 & 22.37 & 154.9 \\

        \rowcolor{oursblue}
        \textbf{Ours}
        & \textbf{51.54\%} & \textbf{16.52} & 0.4623
        & \textbf{0.4704} & \textbf{94.17}
        & \textbf{5.941} & \textbf{11.42} & \textbf{16.14}
        & \textbf{19.23} & \textbf{121.9} \\
        \midrule

        \rowcolor{sectiongray}
        \multicolumn{11}{c}{
            \textbf{Bottom: Evaluation on EgoTeam}
        } \\
        \midrule

        Solaris~\citep{savva2026solaris}
        & 57.65\% & \textbf{16.44} & 0.5673 & 0.4381 & 96.76
        & 0.3162 & 1.247 & 1.422 & 38.91 & 228.9 \\

        $\gamma$-World~\citep{liu2026gammaworld}
        & 52.31\% & 16.18 & 0.5659 & 0.4421 & 97.04
        & 0.3057 & 1.239 & 1.398 & 36.09 & 196.4 \\

        \rowcolor{oursblue}
        \textbf{Ours}
        & \textbf{58.62\%} & 16.33 & \textbf{0.5710}
        & \textbf{0.4373} & \textbf{97.07}
        & \textbf{0.2894} & \textbf{1.232} & \textbf{1.381}
        & \textbf{34.18} & \textbf{195.1} \\
        \bottomrule[0.5pt]
    \end{tabular}%
    }
\end{table*}

\subsection{Comparison with Previous Methods}

\textbf{Quantitative Results.}
We compare~\name{} with Solaris and $\gamma$-World in~\Cref{tab:quantitative_comparison} (top). 
Our method outperforms the baselines on most evaluation metrics in both 4- and 6-player settings, demonstrating strong cross-player consistency, action/camera controllability, and visual quality.
Despite unrestricted cross-player information exchange, Solaris underperforms~\name{} on most metrics while incurring substantial computational overhead, suggesting that dense communication does not necessarily translate into better multi-player generation.
$\gamma$-World struggles with consistency and camera accuracy due to its reliance on the same shared global pathway, which aggregates heterogeneous player information, 
leading to degraded cross-player consistency and player-specific controllability under diverse concurrent interactions.
We attribute~\name{}'s gains to hierarchical communication, which introduces intermediate group-level representations to effectively structure multi-player information exchange.
As pixel-aligned metrics PSNR and SSIM compare pixel values or local image statistics at corresponding spatial locations, they are sensitive to small spatial deviations, 
whereas the consistent gains in GeoKPM and LPIPS indicate stronger geometric and perceptual cross-player consistency of our method.

\textbf{Qualitative Results.}
As shown in~\Cref{fig:qualitative-4_6players},
\name{} demonstrates stronger cross-player consistency, controllability, and interference robustness.
Solaris~\citep{savva2026solaris} fails to preserve consistency across overlapping views
of the same scene or one another, while $\gamma$-World~\citep{liu2026gammaworld}
is more susceptible to information mixing through its shared global pathway, causing inconsistent
visual details and player-specific actions being affected by unrelated observations.
Since all methods share the same video-generation backbone, the results highlight the effectiveness
of hierarchical communication for robust multi-player generation.

\begin{figure}[t]
    \centering
    \includegraphics[width=\linewidth]{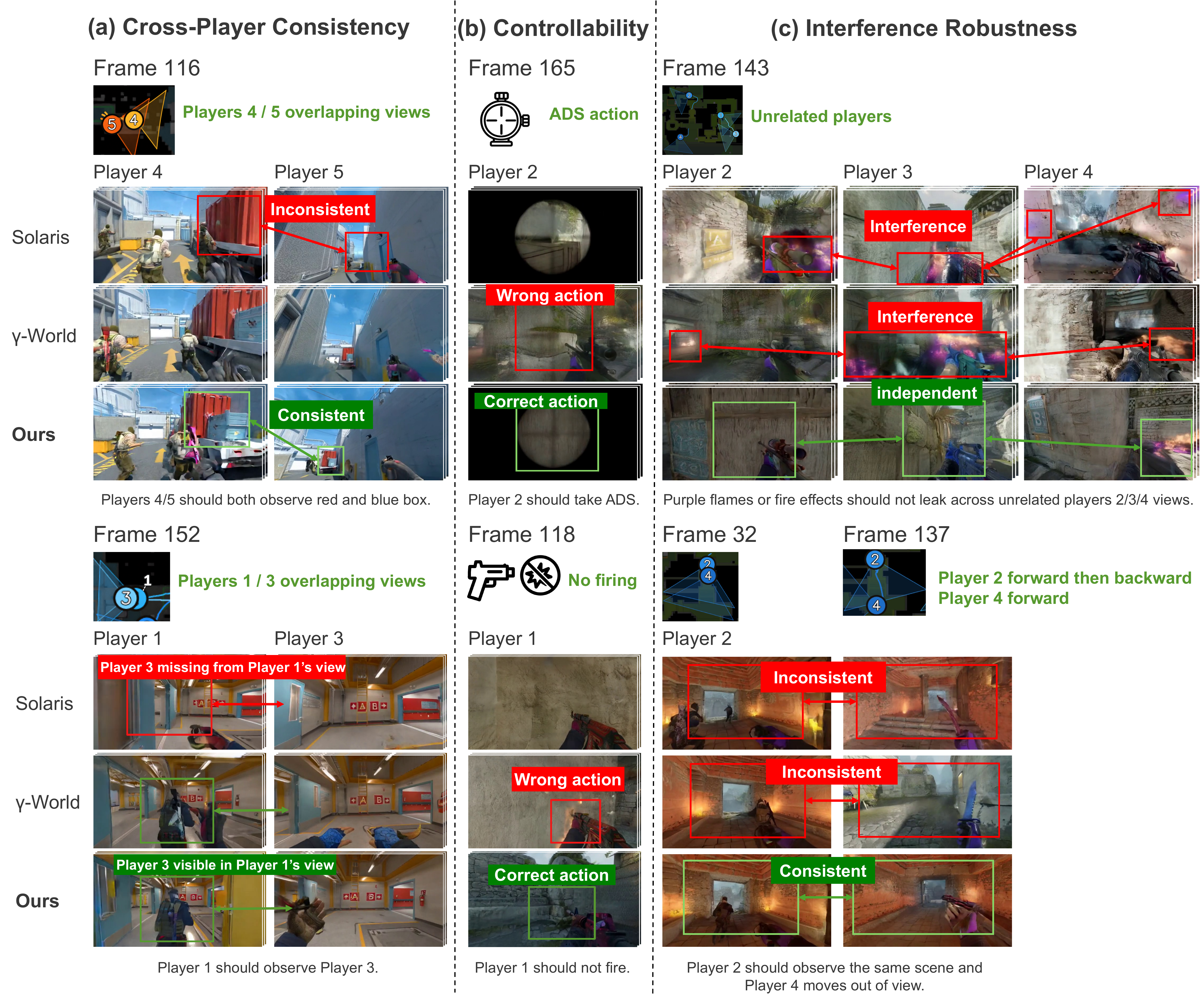}
    \caption{
    \textbf{Qualitative comparison of~\name{} and fully-connected and centralized baselines in 4-/6-player settings.} 
We compare~\name{} with Solaris~\citep{savva2026solaris} and $\gamma$-World~\citep{liu2026gammaworld} on
\textbf{(a)} cross-player consistency: players observing overlapping views of the same scene and one another;
\textbf{(b)} player-specific controllability: adherence to aim-down-sights (ADS) and no-firing controls; and
\textbf{(c)} interference robustness: independence of unrelated concurrent events and preservation of a player's scene upon revisiting despite other players' movements.
Top-down maps are provided for visualization only, and are not used as conditioning inputs.
Other players are omitted for clarity, and can be found in supplementary videos.
    }
    \label{fig:qualitative-4_6players}
\end{figure}

\subsection{Scalability and Generalization Ability of~\name{}}

\textbf{Computational Efficiency.}
Following~\citet{liu2026gammaworld}, we compare communication methods using DiT latency, self-attention latency, and self-attention FLOPs. As shown in~\Cref{fig:efficiency}, fully-connected communication scales quadratically with player count, while~\name{} substantially reduces computation (9.6 vs. 69.1 TFLOPs with 10 players) and remains comparable to centralized and group-based methods. These results demonstrate the computational efficiency of our method for scalable multi-player world models.

\textbf{Generalization to More Player Counts.}
We apply models trained with 4 players to 6-, 8-, and 10-player inference without retraining or introducing additional model parameters.
The results in~\Cref{tab:generalization_egoteam} (middle) highlight~\name{}'s ability to generalize to larger player counts while achieving stronger consistency, controllability, and visual quality than Solaris and $\gamma$-World.

\begin{figure}[t]
    \centering
    \includegraphics[width=\linewidth]{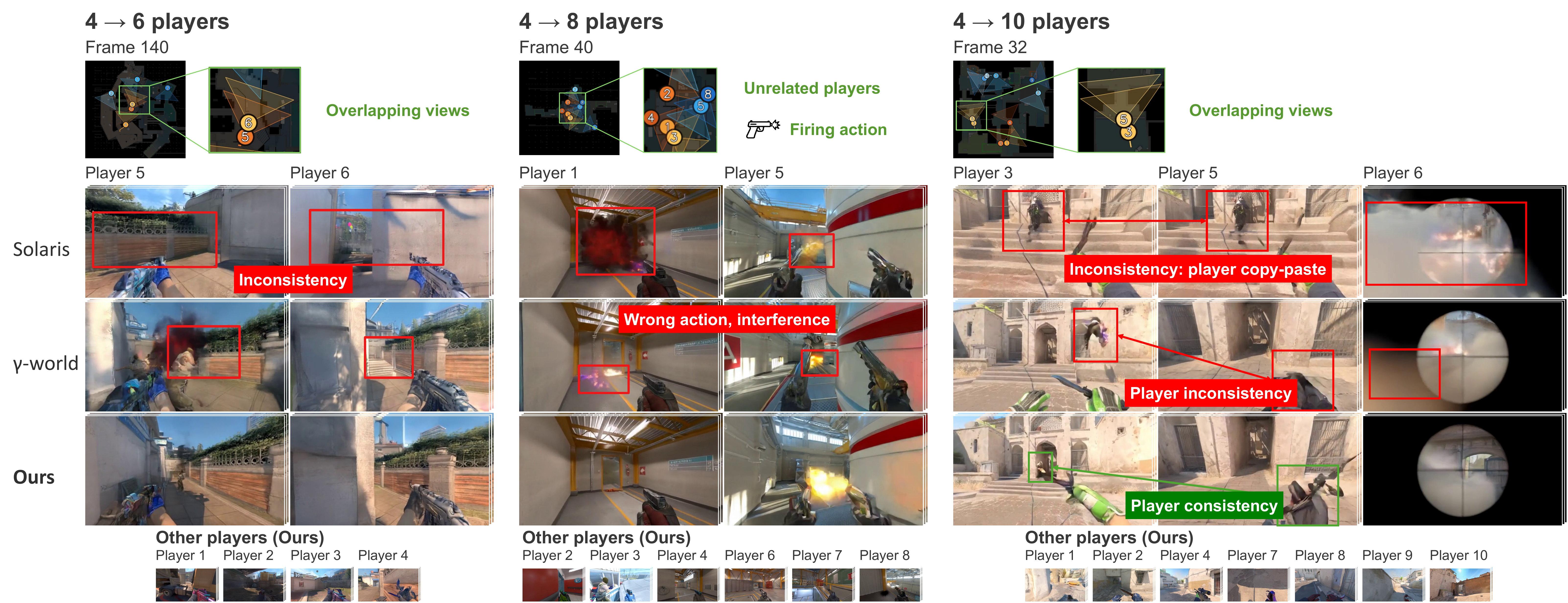}
    \caption{
    \textbf{Qualitative comparison of~\name{} and fully-connected and centralized baselines in
generalizing to 6-/8-/10-player scenarios.}
    }
    \label{fig:4infer_more}
\end{figure}

\textbf{Cross-Domain Applicability.}
Beyond game environments, we also evaluate~\name{} on EgoTeam~\citep{peng2026egoteam}. As shown in~\Cref{tab:generalization_egoteam} (bottom), our method maintains robust multi-player generation performance,
demonstrating its cross-domain applicability.

\definecolor{oursblue}{HTML}{E1EFFA}

\begin{table*}[t]
    \centering
    \caption{\textbf{Ablation on grouping strategies.}}
    \label{tab:grouping_ablation}
    \setlength{\tabcolsep}{5pt}
    \renewcommand{\arraystretch}{1.0}
    \resizebox{\textwidth}{!}{%
    \begin{tabular}{lcccccccccc}
        \toprule
        \multirow{2}{*}{\textbf{Method}}
        & \multicolumn{4}{c}{\textbf{Consistency}}
        & \multicolumn{1}{c}{\textbf{Action Accuracy}}
        & \multicolumn{3}{c}{\textbf{Camera Accuracy}}
        & \multicolumn{2}{c}{\textbf{Visual Quality}} \\
        \cmidrule(lr){2-5}
        \cmidrule(lr){6-6}
        \cmidrule(lr){7-9}
        \cmidrule(lr){10-11}
        & GeoKPM$\uparrow$
        & PSNR$\uparrow$
        & SSIM$\uparrow$
        & LPIPS$\downarrow$
        & IDM$\uparrow$
        & RotErr$\downarrow$
        & TransErr$\downarrow$
        & CamMC$\downarrow$
        & FID$\downarrow$
        & FVD$\downarrow$ \\
        \midrule

        Group-based
        & 48.26\%
        & 17.50
        & 0.4878
        & 0.4317
        & 94.02
        & 2.780
        & 13.18
        & 14.91
        & 28.76
        & \textbf{158.5} \\
        \midrule

        Fixed grouping
        & 50.71\%
        & \textbf{18.16}
        & \textbf{0.5052}
        & 0.4197
        & 94.05
        & 2.496
        & 12.09
        & 13.71
        & 27.78
        & 168.0 \\

        Random grouping
        & 46.76\%
        & 17.78
        & 0.5036
        & 0.4329
        & 93.50
        & 4.796
        & 14.61
        & 17.90
        & 29.83
        & 208.2 \\
        \midrule

        \rowcolor{oursblue}
        \textbf{Ours}
        & \textbf{54.17\%}
        & 18.10
        & 0.5005
        & \textbf{0.4184}
        & \textbf{94.17}
        & \textbf{2.434}
        & \textbf{11.67}
        & \textbf{13.21}
        & \textbf{26.39}
        & 162.3 \\
        \bottomrule
    \end{tabular}%
    }
\end{table*}

\begin{figure}[t]
    \centering
    \includegraphics[width=\linewidth]{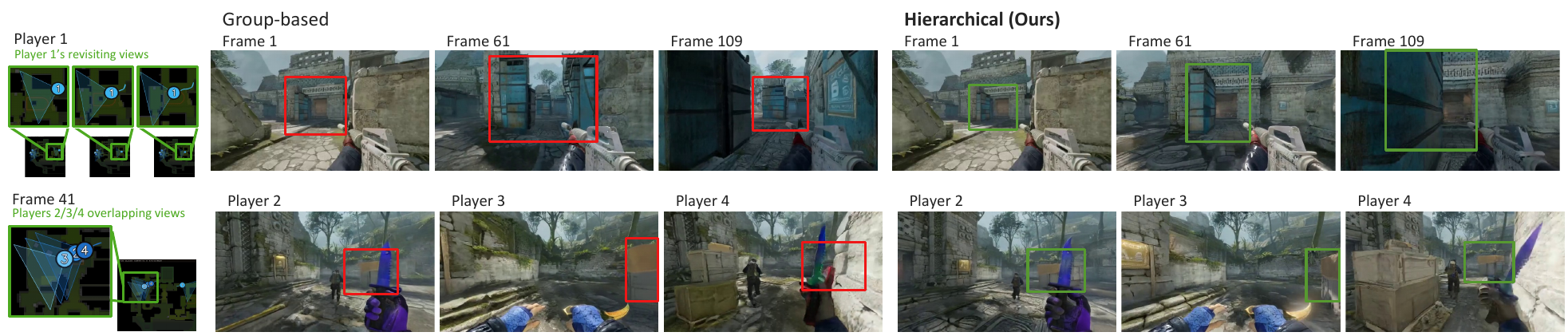}
    \caption{
    \textbf{Qualitative ablation study on communication mechanisms.} The group-based method exhibits inconsistencies when revisiting previously observed scenes (top) and across player views (bottom).
    }
    \label{fig:group-based}
\end{figure}

\begin{figure}[t]
    \centering
    \includegraphics[width=\linewidth]{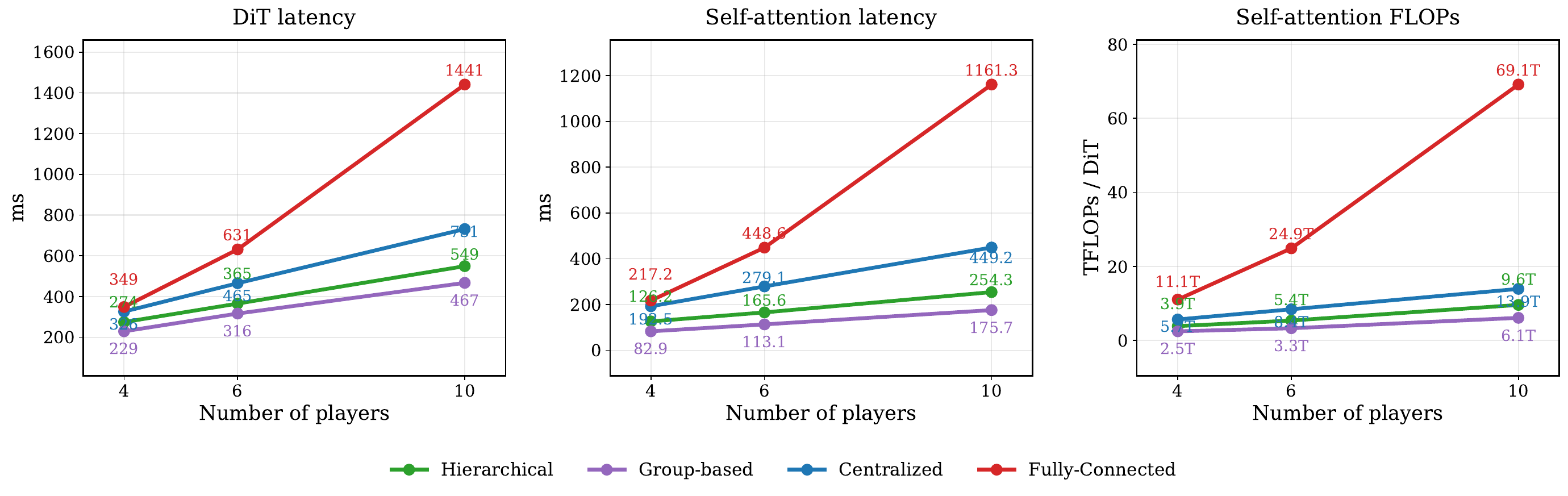}
    \caption{\textbf{Comparisons of computational efficiency.} ~\name{} reduces computation relative to fully-connected communication and achieves efficiency comparable to centralized and group-based methods.
    }
    \label{fig:efficiency}
\end{figure}

\subsection{Ablation Study}

\textbf{Effectiveness of Hierarchical Communication.}
We train four models with different communication schemes on the 4-player OpenCS2 data: fully-connected, centralized, group-based, and hierarchical communication (ours). Results in~\Cref{tab:quantitative_comparison} (top) and~\Cref{tab:grouping_ablation} show that~\name{} consistently improves cross-player consistency and action/camera controllability. In particular, centralized communication is more susceptible to heterogeneous information mixing through its shared global representation, while group-based communication (\Cref{fig:group-based}) restricts local communication but lacks an intermediate group-level representation. These results support the effective use of representations at multiple granularities to structure multi-player information exchange.

\begin{wrapfigure}[12]{r}{0.32\linewidth}
    \vspace{-8pt}
    \centering
    \includegraphics[width=\linewidth]{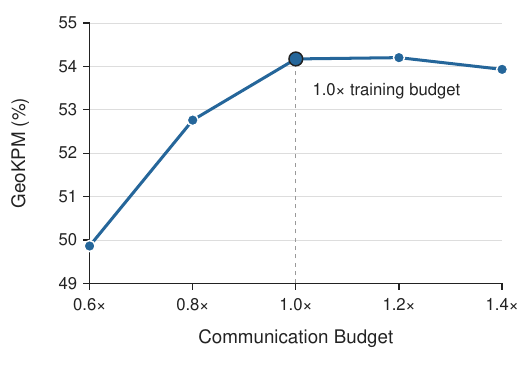}
    \captionsetup{skip=2pt}
    \caption{Effect of communication budget on GeoKPM.}
    \label{fig:geokpm_budget}
    \vspace{-4pt}
\end{wrapfigure}

\textbf{Impact of Dynamic Grouping.}
We train three sets of models with different grouping strategies on 4-player OpenCS2 data:
(1) fixed grouping;
(2) random grouping; 
(3) dynamic grouping (ours).
As shown in~\Cref{tab:grouping_ablation}, fixed grouping yields lower GeoKPM scores and reduced action and camera accuracy compared to dynamic grouping.
Random grouping also leads to substantial decreases in GeoKPM and camera accuracy.
These results demonstrate that dynamic grouping improves consistency and controllability by enabling relevant information exchange among multiple players.

\textbf{Sensitivity to Communication Budget.}
We evaluate how changes in the inference communication budget affect cross-player geometric consistency.
We train the model with the default budget of $1.0\times$ and vary the inference budget from $0.6\times$ to $1.4\times$ without retraining.
As shown in~\Cref{fig:geokpm_budget}, reducing the budget below $1.0\times$ leads to a decrease in GeoKPM, whereas increasing it beyond $1.0\times$ leaves GeoKPM relatively stable.

\FloatBarrier
\section{Conclusion}
We present~\name{}, a multi-player video generation
framework with hierarchical communication that structures information exchange
across player, group, and global levels. 
We introduce group tokens as group-level representations into self-attention, aggregating locally relevant player information while retaining global communication for cross-group exchange. 
We propose shared group-token parameters across group slots, allowing the same communication module to accommodate varying player counts without introducing new group-specific parameters. 
We further introduce budget-constrained dynamic grouping based on recent camera trajectories to determine which players share group-level representations.
Extensive experiments demonstrate~\name{}'s state-of-the-art performance in generating robust multi-player videos with cross-player consistency, controllability, and interference robustness, while scaling to more players.

\endgroup

\bibliography{ref}
\bibliographystyle{iclr2027_conference}

\clearpage
\appendix
\appendix

\section{Implementation Details}\label{appendix:implementation}
We adopt diffusion forcing training~\citep{chen2024diffusionforcing}, with a causal block of 3 latent frames.
We train the model for 10K steps with a local attention window of 7 latent frames, followed by 25K steps with a window of 21 latent frames, using a learning rate of $3 \times 10^{-5}$.
We use AdamW for the two-stage training schedule described above. The per-GPU batch size is one
multi-player sample, and we train on 16 GPUs, resulting in an effective
batch size of 16 multi-player samples. Training is performed in bfloat16
with gradient checkpointing and a gradient clipping norm of 1.0.
We freeze the VAE, text encoder, feed-forward layers, and the remaining
pretrained DiT parameters, while optimizing the self-attention and prompt
cross-attention layers, action control modules, camera Q/K adapters, and
the learnable group and global token embeddings. We apply a dropout rate of
0.1 to both the camera and text conditions.
During inference, we generate 165-frame videos at a resolution of
$352\times640$ using 50 denoising steps and a classifier-free guidance
scale of 5.0.
For hierarchical communication, we use 5 group tokens for each group and 6 global tokens. 
We apply simplex rotary encoding for players~\citep{liu2026gammaworld} to our method and all baselines to make all players permutation-equivariant.

\section{Details of Dynamic Grouping}
\label{appendix:dynamic_grouping}

\textbf{Camera-Based Player Relevance.}
At each grouping step $t$, we estimate pairwise player relevance using only
camera states observed up to the current time, ensuring strict causality.
Given the recent camera trajectories of $N$ players, we compute a symmetric
relevance matrix
\begin{equation}
    \mathbf{R}^{(t)} \in [0,1]^{N\times N},
\end{equation}
where $R^{(t)}_{ij}$ measures the geometric relevance between players $i$
and $j$.

Following the camera-based relevance principle in prior work~\citep{yu2025contextasmemory},
we combine two complementary geometric cues.
First, we estimate \emph{scene relevance} from camera proximity and viewing
direction.
Let $\mathbf{p}_i(\tau)$ and $\mathbf{v}_i(\tau)$ denote the camera position
and unit viewing direction of player $i$ at frame $\tau$.
We compute a proximity score from the camera distance
\begin{equation}
    r^{\mathrm{prox}}_{ij}(\tau)
    =
    \exp\left(
        -\frac{\|\mathbf{p}_i(\tau)-\mathbf{p}_j(\tau)\|_2}{\sigma_p}
    \right),
\end{equation}
and estimate viewing similarity by projecting each camera direction to a
fixed-distance endpoint
$\mathbf{e}_i(\tau)=\mathbf{p}_i(\tau)+\ell\mathbf{v}_i(\tau)$:
\begin{equation}
    r^{\mathrm{view}}_{ij}(\tau)
    =
    \exp\left(
        -\frac{\|\mathbf{e}_i(\tau)-\mathbf{e}_j(\tau)\|_2}{\sigma_v}
    \right).
\end{equation}
Their temporally aggregated combination gives the scene relevance
$R^{(t),\mathrm{scene}}_{ij}$.

Second, we estimate \emph{mutual-view relevance} according to whether the
players fall inside each other's camera fields of view.
Let $Q_{i\rightarrow j}(\tau)\in[0,1]$ denote the pose-based visibility of
player $j$ in player $i$'s view.
We combine the two directed visibility scores as
\begin{equation}
    r^{\mathrm{mutual}}_{ij}(\tau)
    =
    \sqrt{
        Q_{i\rightarrow j}(\tau)
        Q_{j\rightarrow i}(\tau)
    },
\end{equation}
and aggregate it over the recent causal history to obtain
$R^{(t),\mathrm{mutual}}_{ij}$.

We use recency-weighted temporal aggregation and smoothing to make the
relevance robust to short-term camera fluctuations.
The final pairwise relevance is
\begin{equation}
    R^{(t)}_{ij}
    =
    \max\left(
        R^{(t),\mathrm{scene}}_{ij},
        R^{(t),\mathrm{mutual}}_{ij}
    \right).
\end{equation}
Higher values indicate that the corresponding players are more likely to
share locally relevant visual information.
Importantly, the relevance matrix determines only the communication topology;
the actual information exchanged between connected representations is learned
through self-attention.

\textbf{Communication Budget.}
Given $N$ players, we determine the number of available group slots $M$
using a slowly growing function of the player count.
In our implementation, we use
\begin{equation}
    M=\left\lceil\log_2 N\right\rceil.
\end{equation}

We constrain hierarchical communication by the communication cost of the
centralized baseline.
Ignoring the same-player temporal-history term shared by both topologies,
the centralized communication cost is
\begin{equation}
    \mathcal{C}_{\mathrm{central}}
    =
    A_w
    \left(
        2NLk_C + k_C^2
    \right),
\end{equation}
where $L$ denotes the number of visual tokens per player per latent frame,
$k_C$ the number of centralized hub tokens, and $A_w$ the number of allowed
temporal frame pairs.

To derive a preliminary grouping capacity, we use a reference topology
with all local hubs active, each player attached to exactly one local hub,
and no direct player-to-player communication. Its reference cost is
\begin{equation}
    \mathcal{C}_{\mathrm{hier\text{-}base}}
    =
    A_w
    \left(
        2NLk_L
        +
        Mk_L^2
        +
        2Mk_Lk_G
        +
        k_G^2
    \right),
\end{equation}
where $k_L$ and $k_G$ denote the numbers of group and global communication
tokens per latent frame, respectively.
The remaining communication budget is therefore
\begin{equation}
    \Delta\mathcal{C}
    =
    \mathcal{C}_{\mathrm{central}}
    -
    \mathcal{C}_{\mathrm{hier\text{-}base}}.
\end{equation}

Direct intra-group player communication is enabled in all transformer
layers. We express the communication costs on a per-layer basis;
the common number of layers does not affect the budget comparison.
Let $B_b$ denote the number of allowed temporal frame pairs for
direct intra-group player communication.
The resulting preliminary directed player-pair budget is
\begin{equation}
    B_{\mathrm{pair}}
    =
    \left\lfloor
        \frac{\Delta\mathcal{C}}
        {B_b L^2}
    \right\rfloor.
\end{equation}
The candidate group capacity is then bounded by
\begin{equation}
    K
    =
    \min
    \left(
        N,\,
        \left\lfloor
            \frac{1}{2}
            +
            \sqrt{\frac{1}{4}+B_{\mathrm{pair}}}
        \right\rfloor
    \right).
\end{equation}

These preliminary limits restrict the candidate search. For the final
check, $\mathcal{C}(\mathcal{G}^{(t)})$ counts the allowed directed
query--key token pairs in the actual communication mask, excluding
within-player attention shared by all methods. It accounts for overlapping
memberships, global-only players, active and inactive hub attention,
and unique direct player-to-player connections.
We require
\begin{equation}
    \mathcal{C}\bigl(\mathcal{G}^{(t)}\bigr)
    \leq
    \mathcal{C}_{\mathrm{central}},
\end{equation}
ensuring that the final hierarchical topology remains within the centralized
communication budget.
At each grouping step, the search is initialized with an empty set of local
groups, under which all eligible players communicate through the global
tokens. This topology is feasible under the centralized communication budget
in our configuration. Candidate groups are then added only when the resulting
partial topology satisfies all structural and communication constraints;
infeasible candidates are discarded. At the first step, the relevance matrix
is initialized from the camera state observed at $t=0$, so no previous
topology is required.

\textbf{Budget-Constrained Dynamic Grouping.}
Given the relevance matrix $\mathbf{R}^{(t)}$ and the communication
constraints $(M,K)$, we construct the active grouping topology
$\mathcal{G}^{(t)}$.
We first form candidate groups from high-relevance player relations and
retain only candidates that satisfy the group-capacity constraint.
Among feasible candidates, we select at most $M$ groups that prioritize player
coverage while favoring high-relevance player relations and satisfying the
communication budget.

Groups may overlap: each player belongs to at most two groups, and any
two groups share at most one player. A shared player serves as an anchor,
enabling indirect exchange between their local hubs through its visual
representation. The notation $g(i)$ in the main text denotes one group
containing player $i$; the same connection rule applies to each of its
assigned groups.
If unused group slots and budget remain, we attempt to cover unassigned
players with additional feasible groups, reusing an assigned player as
an anchor when needed and subject to the same overlap constraints.
Players that remain uncovered communicate only through the global
communication tokens.

Finally, the resulting variable-size groups are converted into a binary
membership tensor
\begin{equation}
    \mathbf{M}
    \in
    \{0,1\}^{f_z\times M\times N},
\end{equation}
where $M_{t,g,i}=1$ indicates that player $i$ is assigned to group slot $g$
at latent frame $t$.
The membership count $\sum_g M_{t,g,i}\in\{0,1,2\}$ identifies
global-only players, single-group members, and anchors, respectively.
Each player connects to every assigned local hub; players sharing at
least one group have direct attention connections. These connections
form a binary mask and apply in all transformer layers together with
the temporal attention constraints. Memberships may change at each
grouping update.
After converting the selected groups into the time-varying membership tensor,
we additionally evaluate the exact clip-level communication cost induced by
the complete attention mask and require it not to exceed
$\mathcal{C}_{\mathrm{central}}$.

\section{Experimental Results}

\subsection{Additional Results on Generalization to More Player Counts}\label{appendix:player_generalization}
\begin{figure}[t]
    \centering
    \includegraphics[width=\linewidth]{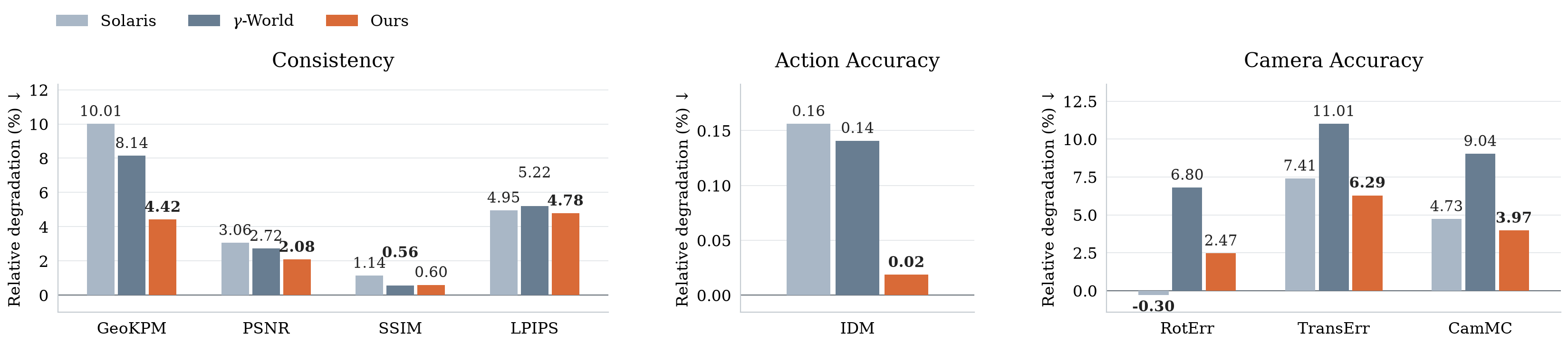}
    \caption{\textbf{Generalization from 4 to 6 players.} Relative performance degradation of models trained with 4 players compared with the same methods trained with 6 players, all evaluated with 6 players. Lower values indicate better preservation of performance; negative values indicate improvement. Bold labels mark the smallest degradation for each metric. Our method achieves the smallest degradation on six of the eight reported metrics.
    }
    \label{fig:generalization}
\end{figure}

To further quantify performance preservation,~\Cref{fig:generalization} reports the relative degradation between $4\rightarrow6$ extrapolation and matched 6-player training.~\name{} exhibits the smallest degradation on most metrics, indicating stronger robustness to unseen player counts.

\begin{table*}[t]
    \centering
    \caption{
        \textbf{Ablation on designs of group-token parameters.}
    }
    \label{tab:ablation_shared_params}
    \setlength{\tabcolsep}{5pt}
    \renewcommand{\arraystretch}{1.15}
    \resizebox{\textwidth}{!}{%
    \begin{tabular}{@{}lcccccccccc@{}}
        \toprule
        \multirow{2}{*}{\textbf{Method}}
        & \multicolumn{4}{c}{\textbf{Consistency}}
        & \multicolumn{1}{c}{\textbf{Action Accuracy}}
        & \multicolumn{3}{c}{\textbf{Camera Accuracy}}
        & \multicolumn{2}{c}{\textbf{Visual Quality}} \\
        \cmidrule(lr){2-5}
        \cmidrule(lr){6-6}
        \cmidrule(lr){7-9}
        \cmidrule(lr){10-11}
        & GeoKPM$\uparrow$
        & PSNR$\uparrow$
        & SSIM$\uparrow$
        & LPIPS$\downarrow$
        & IDM$\uparrow$
        & RotErr$\downarrow$
        & TransErr$\downarrow$
        & CamMC$\downarrow$
        & FID$\downarrow$
        & FVD$\downarrow$ \\
        \midrule

        Copy group 0
        & 45.01\%
        & 16.43
        & 0.4563
        & 0.4797
        & 94.08
        & 4.413
        & 12.91
        & 16.01
        & 26.43
        & 199.0 \\

        Copy group 1
        & 43.63\%
        & 16.41
        & 0.4557
        & 0.4799
        & 94.04
        & 4.606
        & 13.15
        & 16.38
        & 26.39
        & 200.2 \\

        Average both
        & 43.96\%
        & 16.41
        & 0.4557
        & 0.4802
        & 94.04
        & 4.765
        & 13.32
        & 16.62
        & 26.82
        & 202.5 \\

        \textbf{Ours}
        & \textbf{54.33\%}
        & \textbf{16.97}
        & \textbf{0.4659}
        & \textbf{0.4517}
        & \textbf{94.31}
        & \textbf{3.760}
        & \textbf{11.15}
        & \textbf{13.74}
        & \textbf{21.80}
        & \textbf{133.8} \\
        \bottomrule
    \end{tabular}%
    }
\end{table*}

\subsection{Additional Ablation Study}\label{appendix:shared_params}
\textbf{Effectiveness of Shared Group-token Parameters.}
We train two sets of models with different designs for group-token parameters: (1) a model with non-shared group-token parameters, where each group is initialized with an independent set of learnable parameters; (2) our shared group-token model.
Both models are trained on 4-player scenes with two groups and evaluated on 6-player scenes with three groups.
For the unseen third group, we initialize the tokens of the model with non-shared group-token parameters by copying group 0, copying group 1, or averaging both.
As shown in~\Cref{tab:ablation_shared_params}, our design of shared group-token parameters consistently outperforms all non-shared variants across all metrics.
This demonstrates that sharing group-token parameters makes group slots exchangeable, allowing for varying numbers of groups and players, while maintaining robust multi-player generation.

\subsection{More Qualitative Results}
Please see the additional qualitative results in the supplementary video, which has been compressed to 720p due to upload size limitations.

\subsection{Qualitative Results on Cross-Domain Applicability}
\begin{figure}[th]
    \centering
    \includegraphics[width=\linewidth]{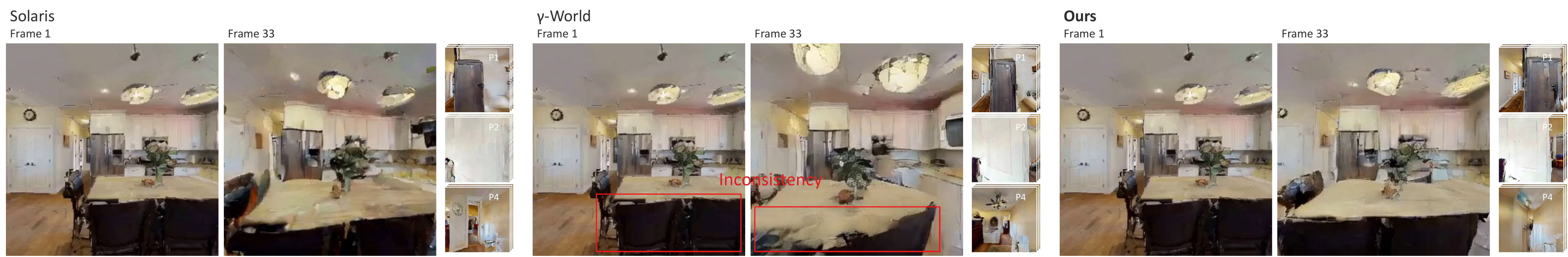}
    \caption{\textbf{Qualitative comparison on EgoTeam.} We compare Solaris, $\gamma$-World, and~\name{} on multi-robot egocentric video generation. 
    }
    \label{fig:egoteam}
\end{figure}

We show qualitative results on EgoTeam~\citep{peng2026egoteam} in~\Cref{fig:egoteam}. EgoTeam contains multi-robot egocentric videos rendered in reconstructed real-world indoor scenes at 256$\times$256 resolution and 5 FPS. 
Compared with $\gamma$-World, our method better preserves cross-view consistency and robustness across multiple agents, demonstrating that the proposed multi-player generation framework is applicable to multi-robot egocentric scenarios beyond gaming environments. 
The quantitative results in~\Cref{tab:generalization_egoteam} also show that our method outperforms the baselines across multiple metrics.

\begin{samepage}
\section{Limitations}
Our current work has several limitations that motivate future research:

1) We primarily focus on cross-player consistency, controllability, and robustness to cross-event interference. Maintaining persistent world states across multiple players over long horizons remains a challenging but important direction for future work.

2) \name{} inherits limitations of the pre-trained video foundation model, including visual artifacts under large and dynamic human motions.
\end{samepage}

\end{document}